\documentclass[10pt,twocolumn,letterpaper]{article}

\usepackage{cvpr}              

\usepackage{graphicx}
\usepackage{amsmath}
\usepackage{amssymb}
\usepackage{booktabs}

\usepackage[pagebackref,breaklinks,colorlinks]{hyperref}

\usepackage[capitalize]{cleveref}
\crefname{section}{Sec.}{Secs.}
\Crefname{section}{Section}{Sections}
\Crefname{table}{Table}{Tables}
\crefname{table}{Tab.}{Tabs.}

\def\cvprPaperID{*****} 
\def\confName{CVPR}
\def\confYear{2023}

\begin{document}

\title{A Car Model Identification System for Streamlining the
Automobile Sales Process}

\author{Said Togru\\
{\tt\small Said.Togru@campus.lmu.de}
\and
\and
Marco Moldovan\\
{\tt\small marco.moldovan@campus.lmu.de}
}

\maketitle

\begin{abstract}
This project presents an automated solution for the efficient identification of car models and makes from images, aimed at streamlining the vehicle listing process on online car-selling platforms. Through a thorough exploration encompassing various efficient network architectures including Convolutional Neural Networks (CNNs), Vision Transformers (ViTs), and hybrid models, we achieved a notable accuracy of 81.97\% employing the EfficientNet (V2 b2) architecture. To refine performance, a combination of strategies, including data augmentation, fine-tuning pretrained models, and extensive hyperparameter tuning, were applied. The trained model offers the potential for automating information extraction, promising enhanced user experiences across car-selling websites.
\end{abstract}

\section{Introduction}
\label{sec:intro}
For individuals looking to sell their cars, online platforms like mobile.de or AutoScout24 offer a convenient way to showcase their vehicles to potential buyers. However, entering all vehicle details can be cumbersome, requiring users to manually input various information such as the car model, manufacturer, etc.  Given that images of the vehicles for sale are typically uploaded in this process, there exists a promising potential for automating the identification and extraction of key details from these images, reducing the chances of errors and expediting the listing process.

Recognizing the need for an efficient and user-friendly solution, our project addresses this challenge by employing a deep learning (DL) model that accurately identifies car models and makes in images. We aim to find and optimize a model architecture that performs well in the given task while still being lightweight and fast.

\section{Theory}
\label{sec:theory}
\subsection{Fine-grained Image Classification}
Fine-grained image classification involves categorizing objects within a broader category into more specific subcategories. Unlike traditional image classification, where the goal is to differentiate between broad object categories (e.g., cats vs. dogs), fine-grained classification targets subtle differences between visually similar subcategories (e.g., different species of birds or, in our case, distinct car models) which makes it more challenging \cite{FineGrai84:online}. 
For example, cars from different manufacturers might share similar body shapes or design cues, creating a potential for misclassification. Additionally, even within a single car model, considerable design differences may exist, such as variations in color, features, accessories, and more. 
\subsection{Deep Neural Networks for Image Classification}
Deep learning has ushered in a paradigm shift in computer vision, revolutionizing the way machines understand visual information. Unlike traditional algorithms that require handcrafted features, deep learning employs neural networks with multiple layers to automatically learn relevant patterns and representations from raw data. This enables the system to decipher intricate details and nuances that are often difficult for human engineers to define.
\subsubsection{Convolutional Neural Networks}
As a subset of deep neural networks, Convolutional Neural Networks (CNNs) occupy a pivotal position in the field of image classification and are therefore ideally suited for our task of recognizing car models and brands from images.

CNNs excel in their ability to perform hierarchical feature extraction. Starting with low-level features in early layers - such as edges and vertices - CNNs progressively move to higher-level features representing complex patterns and objects.
The architecture of CNNs is inspired by the organization of the human visual system. It consists of interconnected layers, each of which serves a specific purpose in the feature extraction process. Convolutional layers use adaptive filters to detect specific features such as edges, textures, and more complex shapes. Subsequent pooling layers downsample the data to preserve essential information while reducing computational overhead. Fully connected layers at the end of the network combine these extracted features to make high-level predictions.

\begin{figure}[htb]
  \centering
  \includegraphics[width=0.5\textwidth]{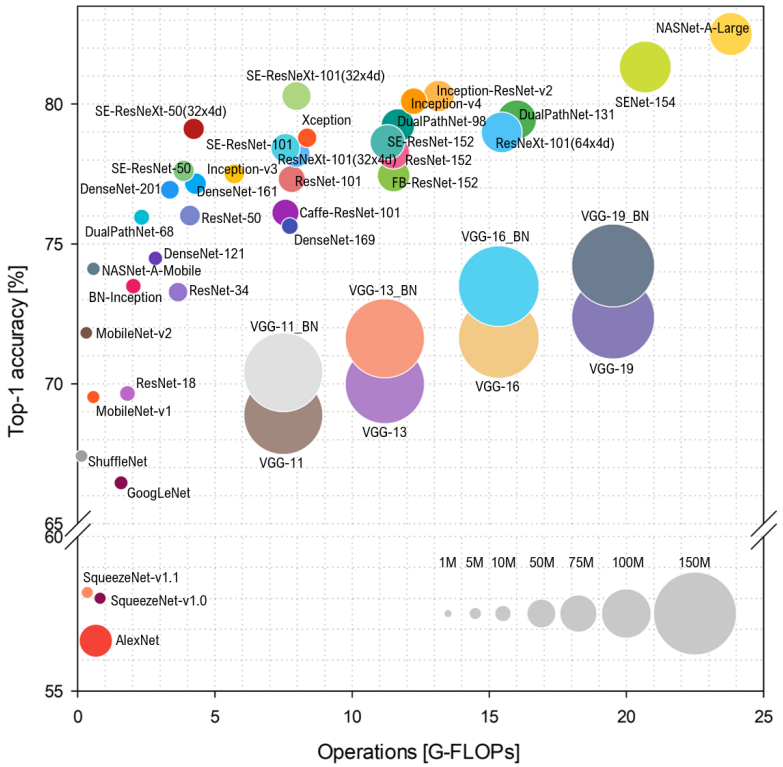}
  \caption{Comparison of ImageNet performances of different model architectures. \cite{Bianco_2018}}
  \label{fig:architectures}
\end{figure}

\begin{figure}[htb]
  \centering
  \includegraphics[width=0.5\textwidth]{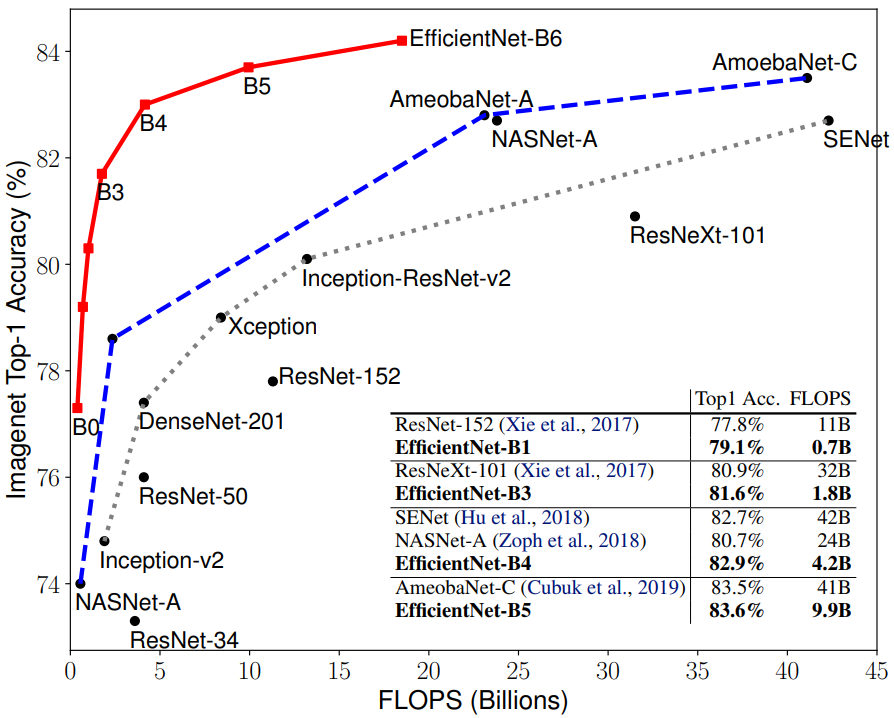}
  \caption{Performance of EfficientNet. \cite{tan2020efficientnet}}
  \label{fig:efficientnet}
\end{figure}

Figure \ref{fig:architectures} provides a comparative overview of various deep neural network architectures in terms of their performances on the ImageNet dataset \cite{Bianco_2018} relative to their computational complexity measured in FLOPs (Floating Point Operations). Certain models stand out for their architectural innovations and remarkable performance. 

One of the most popular CNN architectures \cite{AnOvervi97:online} is ResNet \cite{he2015deep}, which introduced the concept of residual connections. This allowed for the training of much deeper networks than was previously feasible.

In the DenseNet architecture \cite{huang2018densely}, each layer receives input from all preceding layers, improving information flow and reducing the number of parameters. This architecture enhances feature reuse and leads to more compact models. 

EfficientNet \cite{tan2020efficientnet} uses a compound scaling method to scale up the network width, depth, and resolution simultaneously. By optimizing these three dimensions together,state-of-the-art performance is achieved with fewer parameters and lower computational costs (see Figure \ref{fig:efficientnet}). It has become a popular choice for a wide range of computer vision applications due to its efficiency and effectiveness.

Other notable architectures that have made significant contributions to the field of computer vision include Inception \cite{szegedy2015rethinking}, with its multi-scale feature extraction, and Xception \cite{Chollet_2017_CVPR}, with its depthwise separable convolutions. Older popular CNN models, such as AlexNet \cite{NIPS2012_c399862d} and VGG \cite{simonyan2015deep}, have been largely superseded by these more recent architectures. Additionally, smaller-sized models like SqueezeNet \cite{iandola2016squeezenet} and MobileNet \cite{sandler2019mobilenetv2} were designed to prioritize lightweight and efficient computations for mobile deployment, although they do trade off some accuracy to achieve these goals.

\subsubsection{Vision Transformers}
Vision Transformers, or ViTs, offer an alternative to Convolutional Neural Networks (CNNs) for image classification tasks by employing a transformer-based model. While transformers have originally been developed for natural language processing tasks, their application to visual data has yielded remarkable results. For instance, the best-ranked classification models on Papers With Code for the ImageNet dataset used transformer models \cite{ImageNet23:online}. 

The transformer architecture relies on a mechanism known as self-attention, enabling it to capture global dependencies within a sequence of data, whether it's a sentence or an image. In the context of images, the Vision Transformer (ViT) segments the image into fixed-size patches and flattens them into a sequence of vectors. Subsequently, these vectors are processed by the transformer's attention mechanism to capture intricate, long-range relationships between patches. As a result, the model gains the ability to focus on multiple regions within the image, a capability that CNNs may struggle with. This quality makes ViTs particularly intriguing for tasks that demand comprehension of both overarching context and fine-grained distinctions, as in the case of fine-grained image classification. ViTs can outperform CNNs, especially when trained on large datasets. However, ViTs often come with higher computational and memory costs. The self-attention mechanism in transformers has quadratic complexity with respect to the sequence length (number of patches in the case of images), and transformer-based models typically have a large number of parameters, increasing memory requirements and training time.

Recent advancements in transformer architectures have, however, helped reduce the computational costs of ViTs, making them more competitive with CNNs.

MobileViT \cite{mehta2022mobilevit} is a variant of the Vision Transformer designed for mobile and edge devices. It combines the self-attention mechanism of transformers with the efficient convolutional layers of MobileNets. MobileViT is optimized for low computational cost and memory footprint, making it suitable for deployment on resource-constrained devices while still achieving competitive performance on image classification tasks.

Swin Transformer \cite{liu2021swin} is another variant of the Vision Transformer that introduces a hierarchical structure to the self-attention mechanism. It partitions the image into non-overlapping windows and applies self-attention within each window, effectively reducing the computational burden. This local self-attention is then progressively expanded to larger windows in deeper layers, capturing both local and global dependencies.

\subsubsection{Hybrid Models}
Hybrid models in the context of image classification aim to leverage the strengths of both CNNs and transformers, such as the local feature extraction capabilities of CNNs and the global context understanding of transformers. 

One example of a hybrid model is CoAtNet \cite{dai2021coatnet}, which integrates the self-attention mechanism of transformers with the convolutional layers of CNNs. In CoAtNet, convolutional layers are used for feature extraction, capturing local patterns and textures in the image. These extracted features are then fed into transformer layers, which use self-attention to capture global dependencies and relationships between different regions of the image. This combination makes the model versatile and adaptive, suitable for complex visual tasks requiring both detailed feature recognition and broader context understanding.

\section{Data}
\label{sec:data}
To create a robust foundation for our model aimed at streamlining online car sales, we've decided to use an extensive car model dataset. The dataset comprises two key sections: "BMW-10," featuring a set of ultra-fine-grained BMW sedan images (512 images), and "car-197," a comprehensive collection encompassing 197 car models (16,185 images), spanning various categories like sedans, SUVs, coupes, convertibles, pickups, hatchbacks, and station wagons.

The process of dataset construction was not without challenges. Addressing the issue of identifying visually distinct classes, the authors initiated by crawling a prominent car website to obtain a list of car types produced since 1990. A rigorous deduplication approach, employing perceptual hashing, was subsequently applied to example images, resulting in a subset of visually distinct classes, from which we selected 197.

Candidate images were collated from sources like Flickr, Google, and Bing for each class. To ensure effective annotation and a diverse dataset, the candidate images underwent deduplication using the same perceptual hash algorithm, ultimately retaining several thousand images for each of the 197 target classes. These images were then annotated via Amazon Mechanical Turk (AMT) to ascertain their alignment with the respective target classes.

To address the challenge posed by non-expert annotators, we devised a qualification task, complemented by sets of positive and negative example images. In the quest for reliable annotations, we employed the Get Another Label (GAL) system, which estimates candidate image probabilities for target classes and annotator quality.

After compiling images for all 197 classes, we employed AMT to annotate bounding boxes using a quality-controlled system. A final deduplication phase was implemented following the cropping of images to bounding boxes. This carefully curated "Stanford Cars Dataset" serves as a valuable resource, driving the development of a model tailored to optimize online car sales.\cite{6755945}

\section{Implementation}
\label{sec:implementation}

\subsection{Project Architecture}
The successful realization of our optimized 2D car image classification model owes much to the robust tooling and libraries employed in its development. Our implementation leveraged the power and flexibility of PyTorch, a prominent deep learning framework, to orchestrate the core components of our model architecture. PyTorch's intuitive interface and dynamic computation graph facilitated seamless experimentation and efficient model development.\cite{paszke2019pytorch}

To streamline and enhance the training process, we used the PyTorch Lightning scaffold. This abstraction layer not only simplified the training loop implementation but also enabled us to take advantage of Lightning's automated training features, such as distributed training and mixed-precision support. This significantly expedited the model training process and enhanced its scalability.

Configuring and organizing the myriad parameters associated with our model and training pipeline was seamlessly achieved using Hydra. This configuration management tool provided a structured approach to parameter tuning and experimentation, ensuring the reproducibility of results and enabling systematic exploration of hyperparameters.

In addition to model training and configuration management, we enriched our implementation with a user-friendly graphical user interface (GUI) using Gradio. This intuitive library enabled the creation of a user interface for testing and validating the trained models. Through Gradio, stakeholders and users can effortlessly interact with the model, gaining real-time insights into its performance and fine-tuning its behavior.

In summary, our implementation was greatly fortified by the amalgamation of PyTorch for deep learning, PyTorch Lightning for streamlined model training, Hydra for organized configuration, and Gradio for intuitive GUI development. This comprehensive toolset empowered us to not only develop a highly accurate and robust model for 2D car image classification but also facilitated seamless model testing, validation, and integration within the context of online car sales optimization.

\subsection{Data Augmentation}
\subsubsection{Data Preprocessing and Augmentation: Enhancing Model Robustness}

Data augmentation plays a crucial role in bolstering the effectiveness and generalization capabilities of models trained on limited datasets. In our endeavor to optimize online car sales through 2D image classification, we have extensively employed a range of data augmentation techniques to enhance the robustness of our model.

\subsubsection{Importance of Image Augmentation}

Image augmentation serves as a potent strategy to alleviate the challenges posed by limited training data. By artificially diversifying the dataset, augmentation mitigates the risk of overfitting, where a model becomes excessively tailored to the training samples. Additionally, augmentation imparts the model with the ability to discern salient features under varied conditions, thus enabling it to make accurate predictions on unseen real-world data.

\subsubsection{Augmentation Techniques Employed}

\begin{itemize}
    \item \textbf{Horizontal and Vertical Flip:} Mirroring images along both horizontal and vertical axes augments the dataset by providing variations of the same object from different perspectives, thereby improving the model's ability to recognize objects regardless of their orientation.
    
    \item \textbf{Rotation:} Applying rotational transformations introduces variations in object orientation, enabling the model to learn robust features from objects viewed at different angles.
    
    \item \textbf{Greyscale Conversion:} Transforming images to greyscale removes color information, challenging the model to focus solely on shape and texture, thereby enhancing its ability to generalize across different color representations.
    
    \item \textbf{Gaussian Blur:} Introducing controlled blurring simulates real-world variations in image quality, contributing to improved model resilience against noisy or imperfect input data.
    
    \item \textbf{Random Crop:} Extracting random subregions from images diversifies the training set, forcing the model to learn from different object contexts and spatial distributions.
    
    \item \textbf{Normalization:} Scaling pixel values to a common range standardizes input images, ensuring consistent data representation and aiding convergence during training.
    
    \item \textbf{Adjusted Brightness, Contrast, and Saturation:} Manipulating brightness, contrast, and saturation levels introduces variations in lighting and color balance, enabling the model to recognize objects under varying conditions.
\end{itemize}

Notably, we implemented a stochastic data augmentation module that transforms any given data example randomly. In this work, we were heavily inspired by the data augmentation procedures of SimCLR\cite{DBLP:conf/icml/ChenK0H20} and BYOL\cite{grill2020bootstrap} and similar foundational image representation learning papers.\cite{DBLP:conf/icml/ZbontarJMLD21}\cite{baevski2022data2vec}\cite{caron2018deep} We sequentially apply the above mentioned augmentations according to a corresponding random value with probability 0,5 that we sample per augmentation. This enables us to produce a plethora of different image variants, effectively increasing the size of our dataset by a huge margin.

\subsubsection{Advantages of Augmentation}

The implementation of image augmentation within our data preprocessing pipeline carries a multitude of advantages that are pivotal in augmenting the overall efficacy and adaptability of our model for 2D image classification, a key element in optimizing online car sales. This technique, which introduces controlled variations to the training data, serves as a potent tool in addressing the challenges associated with limited datasets and fostering the development of highly capable models.

\textbf{Enhanced Robustness to Variability:} By subjecting our model to augmented data, we effectively simulate a wider spectrum of real-world scenarios and potential input variations. This exposure to diverse data instances imbues the model with an increased resilience to noise, distortions, and imperfections that might arise in actual online car images. As a result, the model becomes more adept at discerning and accommodating variations in lighting conditions, object orientations, and subtle textual details, ensuring that it can make accurate predictions even when presented with non-ideal inputs.

\textbf{Mitigation of Overfitting:} The risk of overfitting, a common challenge when working with limited training data, is considerably reduced through the incorporation of augmentation techniques. Overfitting occurs when a model becomes overly specialized in memorizing training examples and struggles to generalize to new data. Augmentation counteracts this by introducing a broader range of training instances, preventing the model from fixating on specific features present in the limited original data. This, in turn, encourages the model to learn more abstract and transferable features that are applicable across a wider range of instances.

\textbf{Improved Feature Extraction:} Image augmentation compels the model to focus on higher-level and invariant features rather than overly specific details present in the original training set. This transition towards more generalized feature extraction empowers the model to differentiate between essential characteristics of car images, such as distinguishing between different car models based on structural elements, proportions, and prominent design features. Consequently, the model becomes adept at capturing the intrinsic attributes that define each car type, irrespective of the specific instance's visual intricacies.

\textbf{Enriched Diversity in Learning:} A diverse training dataset, achieved through augmentation, propels the model's learning experience beyond the confines of the original data distribution. This diversity primes the model to recognize patterns and attributes across a broader spectrum of contexts, effectively expanding its knowledge base. As a result, the model's predictions become more accurate and adaptable when exposed to new and previously unseen car images, a vital competence in the context of online car sales where variations in image quality, background, and angles are common.

\textbf{ Practical Real-World Applicability:} In the realm of optimizing online car sales, the real-world applicability of the trained model is of paramount importance. Augmentation ensures that the model is well-equipped to handle the inherent heterogeneity of online car images, a scenario where image quality, lighting conditions, and object orientations can significantly vary. By mimicking these real-world challenges during training, augmentation ensures that the model is well-prepared to seamlessly integrate into the online car sales process, contributing to accurate and reliable classification outcomes.

In conclusion, image augmentation emerges as a pivotal technique within our data preprocessing framework, aligning harmoniously with our overarching goal of streamlining online car sales through refined 2D image classification. By enhancing the model's robustness, mitigating overfitting, promoting feature extraction, embracing diversity, and bolstering real-world applicability, augmentation stands as a cornerstone in our pursuit of creating a highly effective and adaptable model for online car sales optimization.\cite{yang2022image}\cite{perez2017effectiveness}\cite{cubuk2018autoaugment}\cite{8388338}

\subsection{Model Implementation}
\subsubsection{Model Selection}

The PyTorch Lightning Template allowed us to easily try out different models for training due to its separation of concerns. This modular approach facilitated experimentation with various architectures and hyperparameters.

We aimed to select a diverse set of models that would allow us to explore how different architectures perform on the car model classification task. We chose the models according to these factors:


\begin{itemize}
    \item Diversity: We wanted to explore a range of architectures, including CNNs, ViTs, and hybrid models, to understand their strengths and weaknesses in the context of car model classification.
    
    \item Efficiency: A primary consideration was the lightweight nature of the models. As we envisioned the application of the car model identification system for a car selling website, it's imperative to have low inference times. The website would likely have many users, and it is important to provide quick and responsive service to avoid making users wait for information. Lightweight models ensure that even with a high volume of user queries, the system remains efficient. 
    Moreover, we had computational constraints due to our limited resources which already excluded certain large and complex models. 
    One of the indicators for inference time is Floating Point Operations (FLOPs), the number of operations required for a single forward pass through the model.
    
    \item Performance: While efficiency was important, we also wanted models that performed well on the task, i.e. a balance between speed and accuracy (e.g. left upper corner in Figure \ref{fig:architectures}). We used the ImageNet benchmark \cite{ImageNet23:online} as an indicator for performance.
\end{itemize}

We finally selected three CNNs (ResNet, EfficientNet and DenseNet), two ViTs (MobileVit, SwinTransformer) and a hybrid model (CoaTNet).
We implemented our models using the torchvision and timm libraries, which provide a wide range of state-of-the-art neural network architectures. The specific model versions were ResNet50 and DenseNet161 from torchvision as well as EfficienNetV2\_B2, MobileVit\_S, Swin\_S3\_tiny and Coat\_lite\_mini from timm.

\subsubsection{Finetuning Models}
Given our midsized dataset, we adopted a transfer learning approach, utilizing pre-trained models as feature extractors and fine-tuning them on our car model dataset. Transfer learning allows us to leverage the knowledge gained from training on large-scale datasets, such as ImageNet, and apply it to our specific task of car model classification, which has a smaller amount of labeled data.

We froze all layers of the pre-trained model, except for the last layer and the classifier. Freezing the layers means that their weights will not be updated during training, and only the weights of the unfrozen layers will be fine-tuned. This approach can be beneficial when working with a small dataset, as it reduces the risk of overfitting by limiting the number of trainable parameters. It also speeds up the training process, as fewer gradients need to be computed. 

The classifier was replaced by a Dropout Layer, followed by a Linear layer. Dropout is a regularization technique used in neural networks to prevent overfitting. Dropout randomly deactivates a fraction of the neurons in a layer at each iteration, effectively "dropping out" those neurons from the network. The fraction of neurons to drop is specified by the dropout rate, a hyperparameter typically set between 0.2 and 0.5. When dropout is applied, the output of the dropped neurons is set to zero, and the remaining neurons must adapt to the absence of those neurons. This prevents the network from relying too heavily on any single neuron and encourages it to learn more robust and generalized features. 
At inference time, dropout is turned off, and all neurons are used for prediction. However, the weights of the neurons are scaled down by the dropout rate to compensate for the increased number of active neurons compared to training. This ensures that the overall magnitude of the network's output remains consistent between training and inference.

\subsection{ Training Infrastructure and Configuration Architecture}
The architecture of our deep learning system has been constructed to facilitate robust and efficient model training, aiming for optimal performance outcomes. At the heart of our infrastructure lies the "defaults" section within our YAML configuration. This foundational element orchestrates a myriad of training components in a hierarchical manner, spanning from the nuanced intricacies of data ingestion to the precision-driven strategies of hyperparameter optimization.

To achieve modularity, we have constructed distinct YAML files, each designed to oversee specific segments such as datasets, model architectures, callback functions, and sophisticated logging mechanisms. This approach not only fosters clarity and manageability but also ensures that each segment functions seamlessly within the overarching training framework.

Our tool of choice for this venture was Optuna, an open-source hyperparameter optimization framework \cite{DBLP}. Central to Optuna's efficiency is the Tree-structured Parzen Estimator (TPE)\cite{watanabe2023treestructured}. Unlike traditional methods that either exhaustively evaluate a fixed space (grid search) or rely on random chance (random search), TPE utilizes Bayesian optimization principles. It forms a probability model based on previous hyperparameter performances, dividing them into better and worse outcomes. For new trials, TPE then selects hyperparameters that maximize the likelihood of favorable results. This probability-driven approach allows TPE to zone in on optimal regions of the hyperparameter space quickly, offering a blend of efficiency and precision. Integrating Optuna, with TPE at its core, ensured our models were trained effectively, making the most of available computational resources.

To streamline the execution phase of model training, distinct shell scripts have been devised for each model. The modular approach guarantees that the unique prerequisites and configurations specific to each model are met without any cross-interference. This strategy provides a level of granularity and command over every individual training process. Moreover, it is worth noting that our methodology is tactically designed to leverage the computing capabilities of LMU.

\subsection{Initial Training and Hyperparameter Space Exploration}
During the formative stages of our research, the initiation of training for all six models was conducted over a span of 10 epochs. Predominantly, our primary objective during this phase was to delve deep into hyperparameter exploration, rather than maximizing model performance per training run. 

For each model, we employed Optuna to conduct 30 distinct trials. The selection of this number was grounded in both theoretical foundations and practical considerations. At the foremost, 30 trials furnish a sufficiently expansive foray into the hyperparameter space, amplifying the probability of zeroing in on a configuration that approaches optimality. Moreover, based on historical data and empirical findings from similar research endeavors, a set of 30 trials emerged as a sweet spot, balancing the depth and breadth of the hyperparameter search with computational prudence.

Our Optuna setup consisted of following parameter selections:

\begin{itemize}
\item \textbf{model.optimizer.target}: The choice between 'torch.optim.Adam' and 'torch.optim.SGD' aims to investigate the impact of different optimization techniques on model convergence and performance. While Adam tends to adapt learning rates for each parameter and might converge faster, SGD is often appreciated for its generalization and has been a staple in deep learning\cite{zhou2021theoretically}.

\item \textbf{model.net.dropout\_value}: The interval of 0.3 to 0.6 was chosen to test the efficacy of dropout in preventing overfitting. By varying dropout values within this range, we aim to identify a sweet spot where the model maintains robustness without sacrificing training accuracy\cite{liu2023dropout}.

\item \textbf{data.batch\_size}: The choices of 32, 64, and 128 cater to the diverse memory requirements of different models. Given that larger batch sizes like 128 are incompatible with models such as 'swintran' and 'mobilevit' on an 8GB GPU, and also problematic for 'densenet', this decision was a direct nod to ensuring efficient GPU utilization without compromising the integrity of the training process.

\item \textbf{model.scheduler.patience}: The integer interval from 5 to 10 gauges the number of epochs with no improvement before adjusting the learning rate. By exploring this range, we aim to strike a balance between allowing models to learn and preventing prolonged stagnation during training.

\item \textbf{model.scheduler.factor}: Spanning from 0.1 to 0.5, this interval seeks the optimal rate by which the learning rate is reduced once a plateau is detected. It empowers the model to adapt its learning curve without making drastic alterations that might jeopardize training stability.

\item \textbf{model.optimizer.weight\_decay}: The interval from 1e-5 to 1e-3 aims to assess the impact of regularization on preventing overfitting. The exploration of this range aids in determining the ideal balance between maintaining model complexity and ensuring generalization.

\item \textbf{model.optimizer.lr}: With a range from 1e-4 to 1e-2, this parameter delineates the learning rate's influence on training dynamics. It's pivotal to pinpoint the optimal learning rate to ensure swift convergence without overshooting minima.
\end{itemize}

Our hyperparameter exploration was architected to serve a dual-fold purpose: maximizing model efficacy and ensuring computational economy. Drawing inspiration from both canonical research methodologies and cutting-edge practices, our endeavor was to discern the optimizer's influence on the training landscape, with an ambition to elucidate any potential transformative shifts that might guide the trajectory of our subsequent training iterations.

\subsection{Final Training with Optimized Hyperparameters}
As we progressed to the decisive phase of our research, our strategy focussed on maximizing the capabilities of our top-performing models based on the insights derived from our comprehensive hyperparameter exploration. Specifically, EfficientNetV2 and Resnet50 emerged as the standout contenders in terms of accuracy performance, making them our models of choice for the extended training of 100 epochs.

This extension in epochs, a considerable leap from our preliminary phase, was pursued with a clear aim: to tap into the highest performance potential of these models. After optimizing hyperparameters in the prior phase, the extended training session allowed each model to converge to its optimal state.

In light of our findings, it was apparent that longer training on models other than EfficientNetV2 and Resnet50 wouldn't be advantageous. Their relative performance was significantly worse, so additional training on these contenders wouldn't yield any benefits.

The Optuna framework, having fulfilled its role in hyperparameter optimization, was set aside during this phase. The optimal hyperparameters, identified earlier, were directly integrated into each model's setup. The models were therefore set up not just on theoretical ideals but on empirical bests.

Our approach was underpinned by two main objectives:

\begin{itemize}
    \item \textbf{Performance Maximization:} Beyond just the hyperparameters, our extended training was designed to allow each model to fine-tune its internal configurations, ensuring optimal prediction outcomes.
    \item \textbf{Operational Efficiency:} Given the computational demands of training advanced neural networks, employing proven hyperparameters reduced the risk of models encountering suboptimal training trajectories. This ensured optimal performance without needless resource expenditure.
\end{itemize}

This final training phase aimed to navigate the vast expanse of potential model performance, leveraging optimized hyperparameters to ensure EfficientNetV2 and Resnet50 operated at their apex. The specific accuracies and losses achieved during this phase will be detailed in the Results Section.

\section{Results}
\subsection{Initial Training with Optuna Hyperparameter Tuning}
\begin{figure}[htb]
  \centering
  \includegraphics[width=0.5\textwidth]{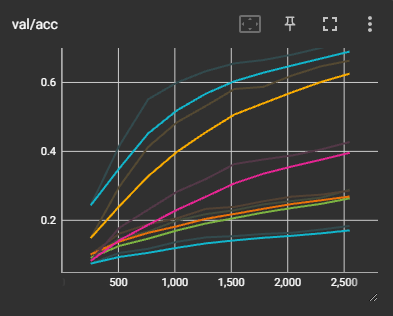}
  \caption{Results from 50 Optuna Trials for each Model}
  \label{fig:optuna}
\end{figure}
Our research commenced with a rigorous hyperparameter tuning using the Optuna framework. Figure \ref{fig:optuna} depitcts the accuracy results for the various models from a run of 50 trials.

The \textit{CoatNet} model achieved an accuracy of 28.60\%, with optimal parameters including a dropout value of 0.304, learning rate of 0.00122, and weight decay of 0.00057.

The \textit{DenseNet161} model posted a superior accuracy of 42.73\% with a dropout value of 0.335, learning rate of 0.000959, and weight decay of 0.00053.

Our star performers, \textit{EfficientNetV2} and \textit{Resnet50}, recorded impressive accuracies of 72.12\% and 66.33\% respectively. Notably, EfficientNetV2 operated best with a dropout of 0.366, learning rate of 0.00157, and weight decay of 0.000216, while Resnet50 favored a dropout value of 0.320, an incredibly low learning rate of 0.000147, and weight decay of 7.91e-05.

On the other end of the spectrum, the \textit{MobileViT} model only reached an accuracy of 18.38\% and \textit{SwinTransformer} posted a result of 28.85\%.

\subsection{Final Training and Validation with Resnet50 and EfficientNetV2}
Having identified the standout models from our initial phase, we proceeded to a more intensive training regimen with Resnet50 and EfficientNetV2, capitalizing on the best parameters discerned from the Optuna trials.
\begin{figure}[htb]
  \centering
  \includegraphics[width=0.5\textwidth]{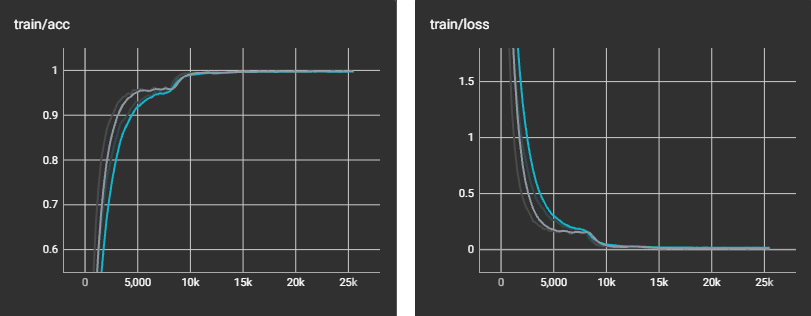}
  \caption{Training Results EfficientNet \& Resnet50}
  \label{fig:train}
\end{figure}
Figure \ref{fig:train} illustrates the training results for both models, emphasizing accuracy and loss graphs over the span of 100 epochs. By the end of this training phase, EfficientNetV2 reported a training accuracy of 99.79\% with a minuscule loss value of 0.00889, while Resnet50's training metrics echoed a similar success story with an accuracy of 99.74\% and a loss of 0.01725.
\begin{figure}[htb]
  \centering
  \includegraphics[width=0.5\textwidth]{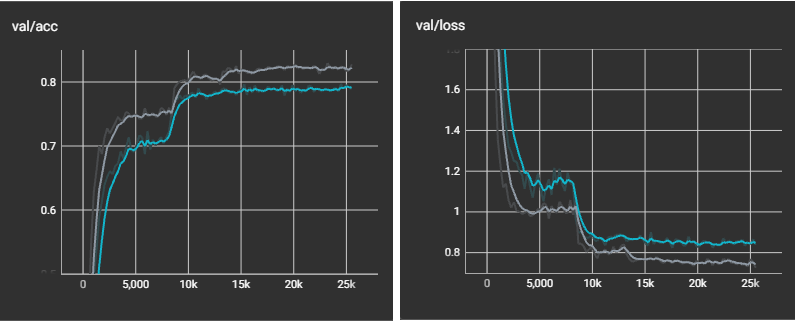}
  \caption{Validation Results EfficientNet \& Resnet50}
  \label{fig:val}
\end{figure}
Figure \ref{fig:val} shows the validation results for both models. EfficientNetV2 managed a validation accuracy of 82.74\% against a loss value of 0.7258. Resnet50, in its stride, recorded a validation accuracy of 78.91\% with a loss figure of 0.8385.

To further test the efficacy of our models, a separate test phase was conducted. EfficientNetV2 achieved a compelling test accuracy of 81.97\%, outpacing Resnet50 which secured a commendable 77.81

\section{Discussion}
\subsection{Findings}
Our training yielded several insightful findings, touching upon model performance, the substantial impact of hyperparameters, and potential overfitting concerns:

\begin{itemize}
\item \textbf{Model Performance Differentiation:} As visualized in Figure \ref{fig:architectures}, there's a pronounced variance in the accuracy and efficiency metrics among the models. While \textit{EfficientNetV2} and \textit{Resnet50} stood out as frontrunners with high accuracies, other models, notably \textit{MobileViT} and \textit{CoatNet}, lagged considerably behind. Such disparities highlight the intrinsic differences in the architectures and their adaptability to our specific dataset. The comparatively lower performance of the transformer models could be attributed to certain inherent challenges of Vision Transformers. ViTs often require larger and more diverse datasets to effectively capture global context through self-attention mechanisms. Our dataset's limited size might have hindered their ability to learn optimal features. Moreover, the self-attention mechanism processes all pairs of positions, making the model inherently more resource-intensive compared to Convolutional Neural Networks (CNNs) which is why we chose smaller versions of the architectures. Additionally, ViTs tend to have a higher number of parameters, potentially leading to overfitting in cases with small datasets.

\item \textbf{Significance of Hyperparameters:} Our hyperparameter search showed the significance of choosing the correct hyperparameters, as the performances of the trials had a large variance. The commendable uplift in performance for the models is largely attributed to their optimized hyperparameters, underscoring the efficacy of the Optuna framework.

\item \textbf{Peak Performances in Final Training:} The data from Figure \ref{fig:train} and Figure \ref{fig:val} demonstrates the capabilities of \textit{EfficientNetV2} and \textit{Resnet50}. Both models achieved near-perfect training accuracies, hinting at a possible overfitting scenario. The high training accuracies, nearing 100\%, juxtaposed against the validation scores of around 80\% suggest that while the models have learned the training data exceptionally well, they might be slightly over-optimized for it, potentially reducing their generalizability to new, unseen data.

\item \textbf{Test Accuracies:} The test accuracies provide an additional lens to gauge model performance on unencountered data. \textit{EfficientNetV2} showed an impressive accuracy of 81.97\%, closely trailed by \textit{Resnet50} at 77.81\%. These figures validate the robustness of our trained models but also leave room for improvement, especially when compared to the training accuracies.
\end{itemize}

In essence, our findings emphasize not just the different performances of different neural network architectures but also spotlight the influence of methodical hyperparameter tuning on model performance and the ever-present challenge of ensuring models remain generalizable across diverse datasets.
\subsection{Limitations}

In the course of this research, while we aimed to maintain high standards of scientific and technical precision, certain limitations inevitably influenced the outcomes of our work. These limitations, largely due to time constraints and hardware challenges, played a significant role in our research decisions and methodologies.

\begin{itemize}
    \item \textit{GPU Limitations}: Our research encountered specific challenges concerning the GPU capacity, which posed constraints on batch sizes. For instance, the `128` batch size was infeasible for models like 'swintran' and 'mobilevit', owing to GPU capacity constraints. Similarly, both `128` \& `64` batch sizes proved incompatible with 'densenet'. These limitations underscore the role of hardware in determining the scalability and efficiency of deep learning models.
    \item \textit{Computational Resources}: The computing resources provided by LMU were not always sufficient to cater to the intense demands of our experiments. With each Optuna trial ranging between 30 minutes to 1 hour, contingent upon the 10 epoch duration and the chosen hyperparameters, it became evident that a richer resource pool would have expedited our experimentation phase.
    \item \textit{Throttling of Resources}: During the course of our experiments, we encountered unexpected throttling of computational resources. The origins of this throttling remain unclear - it could potentially be an artifact of cluster and server policies. Such throttling poses challenges, especially when striving for consistent performance benchmarks.
    \item \textit{Server Restart Issues}: A challenge that hindered our progress was the unanticipated server restarts at 7 am. Encountering a restart during the midst of an Optuna hyperparameter search not only disrupted the continuity of the research but also necessitated a complete restart. Resuming an interrupted Optuna trial without prior preparation for such eventualities is intricate, leading to lost computational time and resources.
\end{itemize}

In summary, while our research was ambitious in its scope and intent, it was not without its set of challenges. These limitations, both anticipated and unexpected, underscore the nature of deep learning research, where scientific ambition often intersects with real-world logistical and hardware constraints. It serves as a reminder of the importance of contingency planning and adaptability in the rapidly evolving landscape of AI research.

\section{Conclusion}
By using deep neural networks, we accomplished to accurately and efficiently identify car models and makes from images with a validation accuracy of  81.97\%. Among the diverse set of model architectures we explore, we selected EfficientNet (V2 b2) as our final candidate for this task. It outperformed other CNNs, ViTs and hybrid architectures in terms of accuracy, speed and efficiency. We implemented a range of strategies to optimize our network's performance. We applied data augmentation to enhance the model's robustness, used fine-tuned pretrained models and conducted an extensive hyperparameter search.

Our trained model can be used in the listing process on car selling platforms to automatically fill in information retrieved from the images. For future developments, it is conceivable to extract further information from the images, such as color, number of doors and supplementary features to further improve user experience on car selling websites. Moreover, other versions of EfficientNet can be explored, coupled with varying degrees of layer freezing to fine-tune performance even further.

While we achieved commendable results with our CNN network, we recognize that the field of Vision Transformers holds promise. With their emergence as a novel technology, it's conceivable that more refined efficient models will evolve, potentially offering stiff competition to the established CNNs.

In summary, our project underscores the capabilities of deep neural networks in solving fine-grained image classification tasks with notable accuracy and manageable computational demands.

{\small
\bibliography{Paper}
\bibliographystyle{ieee_fullname}
}

\end{document}